\documentclass[]{article}
\usepackage{proceed2e}
\usepackage{subfigure}
\usepackage{graphicx} \usepackage{amsmath} \usepackage{amssymb}
\usepackage{epic}\usepackage{eepic}
\newcommand{\BEQ}{\begin{equation}}
\newcommand{\EEQ}{\end{equation}}
\newcommand{\BEA}{\begin{eqnarray}}
\newcommand{\EEA}{\end{eqnarray}}
\newcommand{\BGA}{\begin{gather}}
\newcommand{\EGA}{\end{gather}}

\newcommand{\e}{\epsilon}

\newcommand{\half}{\frac{1}{2}}

\newcommand{\bx}{{\bf x}}
\newcommand{\by}{{\bf y}}

\newcommand{\bp}{{\bf p}}
\renewcommand{\o}{\frac{1}{2}}
\newcommand{\oo}{\frac{1}{4}}

\newcommand{\w}{\omega}
\newcommand{\comment}[1]{}
\title{On Maximum a Posteriori Estimation of Hidden Markov Processes} 

\author{ {\bf Armen Allahverdyan 
} 
\\
Yerevan Physics Institute \\  
Yerevan 375036, Armenia \\ 
aarmen@mail.yerphi.am
\And 
{\bf Aram Galstyan}  \\ 
USC Information Sciences Institute\\
Marina del Rey, CA 90292, USA \\    
galstyan@isi.edu         
}

\begin{document} 
 
\maketitle 
 
\begin{abstract} We present a theoretical analysis of Maximum a
Posteriori (MAP) sequence estimation for 
binary symmetric hidden Markov processes. We reduce the MAP
estimation  to the energy minimization of an 
appropriately defined Ising spin model, and focus on the performance of MAP as
characterized by its accuracy and the number of solutions
corresponding to a typical observed sequence.  It is shown that for a
finite range of sufficiently low noise levels, the solution is uniquely
related to the observed sequence, while the accuracy degrades linearly
with increasing the noise strength. For intermediate noise values, the
accuracy is nearly noise-independent, but now there are {\em exponentially} many
solutions to the estimation problem, which is reflected in non--zero ground--state entropy
for  the Ising model. Finally, for even larger noise intensities, the number of solutions reduces again, but the
accuracy is poor. It is shown that these regimes are different thermodynamic phases of
the Ising model that are related to each other via first-order phase transitions.
\end{abstract}
 
\section{Introduction} 

Hidden Markov Models (HMM) are used extensively for modeling sequential data
in various areas \cite{rabiner_review,ephraim_review}: information
theory, signal processing, bioinformatics, mathematical economics,
linguistics, {\it etc}.  One of the main problems underlying many applications of
HMMs amounts to inferring the hidden state sequence $\bx$ based on
noise-corrupted observation sequence $\by$. This is often done through  maximum a posteriori (MAP) approach, which finds an estimate $\hat{\bx}(\by)$ by maximizing the posterior probability ${\rm Pr}(\bx|\by)$.

The computational solution to the MAP optimization problem is
readily available via the Viterbi algorithm \cite{rabiner_review}. Despite its extensive use  in many applications, however,  the properties of MAP estimation, and specifically, the structure of its solution space, have received surprisingly little attention. On the other hand, it is clear that choosing a single state sequence might be insufficient for adequately understanding the structure of the inferred process. To get a more complete picture, one needs to know whether there are other nearly optimal sequences, how many of them, how they compare with the optimal solution, and so on.

Generally, the structure of an inference method can be characterized by the accuracy of the estimation, 
and the number ${\cal N}(\by)$ of solutions $\hat{\bx}(\by)$ that the method can produce in
response to a given sequence $\by$. In this paper we study the structure of MAP inference for the simplest binary,
symmetric HMM. As an accuracy measure we employ the moments of the
estimated sequence $\hat{\bx}$ in comparison of those of the actual
sequence $\bx$, while the number ${\cal N}(\by)$ of possible estimates
will be characterized by its averaged logarithm ${\sum}_\by{\rm
Pr}(\by)\ln {\cal N}(\by)$.  The binary symmetric HMM is studied by
reducing it to the  Ising model in random fields, a relation well-known
both in computer science \cite{geman} and statistical physics
\cite{scot}. In this way, the average cost $-{\sum}_\by{\rm Pr}(\by){\rm
Pr}(\hat{\bx}(\by)|\by)$ of MAP and the logarithm of the number of solutions
${\sum}_\by{\rm Pr}(\by)\ln {\cal N}(\by)$ relate, respectively, to
the energy and the entropy of the Ising model at the zero temperature.

Our results indicate that even for  a simple process such as binary symmetric HMM,  MAP yields a very rich and non-trivial   solution structure. The main findings can be summarized as follows:  For a small, but finite range of noise values the MAP solution is uniquely related to the observed sequence, and the accuracy of the solution degrades linearly
with increasing the noise strength.  For intermediate values of noise the accuracy is nearly noise-independent, but now there are {\em exponentially} many
solutions to the estimation problem, which is reflected in non--zero ground--state entropy
for  the Ising model. Finally, for  larger noise intensities the number of solutions is reduced again, but the
accuracy is poor. Furthermore,  those regimes are the manifestation of different thermodynamic phases of
the Ising model, which are related to each other via first-order phase transitions.

The rest of the paper is organized as follows:  After some general discussion of MAP scheme in Section \ref{map}, 
we define the model studied here in Section~\ref{bshmm}. Its solution is given in Sections
\ref{recursion} and \ref{mapo}. The latter also discusses our concrete
findings on the structure of MAP for the  binary symmetric HMM. We conclude the paper by discussion of our results and future work. 

\section{Maximum a posteriori (MAP) estimation: general description}
\label{map}
Let $\bx=(x_1,\ldots,x_N)$ and $\by=(y_1,\ldots,y_N)$ be realizations of
discrete-time random processes ${\cal X}$ and ${\cal Y}$, respectively.
We write their probabilities as $\bp (\bx)$ and $\bp (\by)$.  We assume that
${\cal Y}$ is the noisy observation of ${\cal X}$; the influence of
noise is described by the conditional probability $\bp(\by
|\bx )$. Let us further assume that we are given an observed sequence $\by$, and we know the
probabilities $\bp(\by |\bx )$ and $\bp(\bx )$. We do not know which
specific sequence $\bx$ generated the observation $\by$. MAP offers a
method for estimating the generating sequence $\hat{\bx} (\by)$ on the
ground of $\by$: $\hat{\bx}$ is found by maximizing over $\bx$ the
posterior probability $\bp(\bx|\by)={\bp(\by|\bx)\bp(\bx)}/
{\bp(\by)}$.  Since ${\bp(\by)}$ does not depend on $\bx$, we can
equally well minimize
\BEA
\label{2}
-\ln [\, \bp(\by|\bx)\bp(\bx)\,] \equiv H(\by,\bx).
\EEA
The advantage of using $H(\by,\bx)$ is that if ${\cal Y}$ is ergodic (in the sense of weak law of large numbers)
\cite{cover_thomas}, which we assume from now on, then for $N\gg 1$,
$H(\by,\hat{\bx} (\by))$ will be independent from $\by$, if $\by$
belongs to the typical set $\Omega_N ({\cal Y})$ \cite{cover_thomas}.
The typical set has the overall probability converging to one:
$\sum_{\by \in \Omega_N ({\cal Y}) }\bp (\by)\to 1$. Since all elements
of $\Omega_N ({\cal Y})$ have (nearly) equal probability, we can employ
with probability one the averaged quantity $\sum_{\by}\bp (\by)
H(\by,\hat{\bx} (\by))$ instead of $H(\by,\hat{\bx} (\by))$. 

If the noise is very weak, $\bp(\bx|\by)\simeq \delta (\bx-\by)
=\prod_{k=1}^N \delta(x_k-y_k)$ (with $\delta(x)$ being the Kronecker
delta), we recover the generating sequence almost exactly. For a strong
noise the estimation is dominated by the prior distribution
$\bp(\bx|\by)\simeq \bp(\bx)$, so that the estimation is not informative.
When no priors are put, $\bp (\bx)\propto {\rm const}$, the MAP
estimation reduces to the Maximum Likelihood (ML) estimation scheme.
The latter also reproduces the  source sequence almost exactly if the
noise is weak. 

According to the Viterbi algortithm, for a given $\by$ 
the mimimization of $H(\by,\bx)$ in (\ref{2}) produces
one single estimate $\hat{\bx}(\by)$. However, it is possible
that there are other sequences $\hat{\bx} ^{[\alpha]} (\by)$ for which 
$H(\by, \hat{\bx} ^{[\alpha]} (\by) ) $, though greater than 
$H(\by, \hat{\bx}  (\by) ) $, is almost equal to the latter in the 
sense of 
${\rm lim}_{N\to\infty} \frac{H(\by, \hat{\bx} ^{[\alpha]} (\by) )]}{N}={\rm lim}_{N\to\infty}
\frac{H(\by, \hat{\bx}  (\by) )}{N} $. All such sequences are equivalent for $N\to\infty$ and we
list them as possible solutions:
\BEA
\label{gumilev}
\hat{\bx} ^{[\alpha]} (\by), \qquad \alpha = 1,\ldots, {\cal N}(\by).
\EEA
If $\ln {\cal N}(\by) \propto N$, 
we repeat the above ergodicity argument and get
for the logarithm of the number of solutions corresponding 
to a typical observed sequence
\BEA
\label{campo_pretorius}
\Theta= {\sum}_{\by}\bp (\by) \ln {\cal N}(\by).
\EEA
A finite $\frac{\Theta}{N}$ means that there are exponentially many
outcomes of minimizing $H(\by,\bx)$ over $\bx$. We call $\Theta$
entropy, since it relates to the entropy of the Ising model; see below. 

We can calculate various moments of $\hat{\bx} ^{[\alpha]} (\by)$, which
are random variables due to the dependence on $\by$, and employ them for
characterizing the accuracy of the estimation; see below for 
examples. For small noise values these moments will be close to those of the
original process ${\cal X}$.  Another useful quantity is the average
overlap between the estimated sequences $\hat{\bx} ^{[\alpha]} (\by)$,
and the observed sequence $\by$ (definition of overlap is clarified
below). A small overlap means that the estimation is not dominated by
observations.

\section{Binary symmetric hidden markov model (HMM).}
\label{bshmm}

\subsection{Definition.}

We consider the MAP estimation of a binary, discrete-time Markov stochastic process ${\cal X}=(X_1,X_2,\ldots,X_N)$. 
Each random variable $X_k$ has only two realizations $x_k=\pm 1$. The Markov feature implies
\BEA
\label{orleans}
\bp(\bx)={\prod}_{k=2}^N p(x_{k}|x_{k-1})p(x_1),
\EEA
where $p(x_{x}|x_{k-1})$ is a time-independent transition probability of the Markov process. For the considered binary symmetric situation it is parameterized by a single number $0<q<1$,  $p(1|1)=p(-1|-1)=1-q$, $p(1|-1)=p(-1|1)=q$, and the
stationary distribution is $p_{\rm st}(1)=p_{\rm st}(-1)=\frac{1}{2}$. Furthermore, the noise process is assumed to be memory-less, time-independent and unbiased:
\begin{gather}
\label{bejrut}
\bp(\by|\bx) = {\prod}_{k=1}^N \pi (y_k|x_k), \qquad y_k=\pm 1
\end{gather}
where  $\pi(-1|1)=\pi(1|-1)=\epsilon$, $\pi(1|1)=\pi(-1|-1)=1-\epsilon$, and  $\epsilon$ is the probability of error. Here memory-less
refers to the factorization in (\ref{bejrut}), time-independence refers
to the fact that in (\ref{bejrut}) $\pi (...|...)$ does not depend
on $k$, while unbiased means that the noise acts
symmetrically on both realizations of the Markov process: 
$\pi(1|-1)=\pi(-1|1)$.

Note that the composite process
${\cal X}{\cal Y}$ with realizations $(y_k,x_k)$ is Markov with transition probabilities
\BEA
\label{epatage}
p(y_{k+1},x_{k+1}| y_{k}, x_{k} )=\pi(y_{k+1}|x_{k+1}) p( x_{k+1}| x_{k} ).
\EEA
However, ${\cal Y}$ is in general not a Markov process. 

\subsection{Mapping to the Ising model.}

Let us represent the transition probabilities as
\begin{gather}
\label{om}
p(x_{k}|x_{k-1})=\frac{e^{J x_{k} x_{k-1} }}{2\cosh J} , ~~~~
J=\frac{1}{2}\ln\left[\frac{1-q}{q } \right].
\end{gather}
Likewise, we represent the noise model  as
\BEA
\pi (y_{i}|x_{i})=\frac{e^{h y_i x_i}}{2\cosh h}, ~~~~
h=\frac{1}{2}\ln \left[
\frac{1-\epsilon}{\epsilon}
\right].
\label{kaskad}
\EEA
We combine (\ref{2}, \ref{orleans}--\ref{bejrut}) to represent the log--likelihood as
\BEA
\label{jershalaim}
H(\by,\bx)=-J{\sum}_{k=1}^N
x_{k} x_{k+1} 
-h{\sum}_{k=1}^Ny_kx_k,
\EEA
where we have omitted an irrelevant additive factor.
$H(\by,\bx)$ is the Hamiltonian of a  one--dimensional (1d) Ising spin model with external
random fields $hy_k$ governed by the
probability $\bp(\by)$ \cite{zuk}. The factor $J$ in
(\ref{jershalaim}) is the spin--spin interaction constant, uniquely determined from the transition 
probability $q$: If $q<\frac{1}{2}$, the constant $J$ is
positive, which refers to the {\em ferromagnetic} situation: the spin--spin
interaction tends to align the spins. From now on we assume $J>0, h>0$.
\comment{\BEA
\label{paris}
J>0,\qquad h>0.
\EEA
}
We note that the main difference between (\ref{jershalaim}) and other random--field Ising
models considered in literature \cite{scot,behn}, is that in our
situation the random fields are not uncorrelated random variables, but
display non-Markovian correlations. 

\subsection{Implementation of MAP}

To minimize $\sum_{\by} \bp (\by) H(\by,\bx)$
over $\bx$,  we introduce a
non-zero temperature $T=\frac{1}{\beta}\geq 0$, and define the following
conditional probability
\BEA
\rho (\bx|\by)\equiv 
\frac{e^{-\beta H(\by,\bx)}}{Z(\by)},\qquad 
Z(\by)\equiv {\sum}_{\bx} e^{-\beta H(\by,\bx)},
\label{bala}
\EEA
where $Z(\by)$ is the partition function.  In the terminology of
statistical physics, $\rho (\bx|\by)$ gives the probability distribution
of states $\bx$ for a system with Hamiltonian $H(\by,\bx)$ interacting
with a thermal bath at temperature $T$, and with frozen (i.e., fixed for
each site) random fields $y_k$ \cite{landau}.  For $T\to 0$, and a given
$\by$, the function $e^{-\beta H(\by,\bx)}$ is strongly picked at those
$\hat{\bx}(\by)$ [ground states], which minimize $H(\by,\bx)$. If,
however, the limit $T\to 0$ is taken after the limit $N\to\infty$, we get
\BEA
\rho (\bx|\by)\to \frac{1}{{\cal N}(\by)}\sum_{\alpha}
\delta [\bx - \hat{\bx}^{[\alpha]}(\by) ],
\label{chandr}
\EEA
where $\hat{\bx}^{[\alpha]}$ and ${\cal N}(\by)$ were defined in (\ref{gumilev}). From now on
we understand the limit $T\to 0$ in this sense.

The average of $H$ [average energy] in the $T\to 0$ limit will be equal to the $H(\by,\bx)$ minimized over
$\bx$:
\BEA
{\sum}_{\bx \by} \bp(\by)\rho (\bx|\by) H(\by, \bx)= {\sum}_{\by } \bp(\by) H(\by, \hat{\bx}^{[1]}(\by)).
\label{palatin}
\EEA
where we have used the fact that all ground state configurations $\hat{\bx}(\by)$  
have the same energy, $H(\by, \hat{\bx}^{[\alpha]})=H(\by, \hat{\bx}^{[1]})$, for any $\alpha$.  

The average logarithm $\Theta$ of the number of 
MAP solutions is equal to the zero-temperature entropy
\BEA
\Theta=
-{\sum}_{\bx\by} \bp(\by)\rho (\bx|\by) \ln \rho (\bx|\by)=
{\sum}_{\by} \bp(\by)\ln {\cal N}(\by). \nonumber
\EEA

Let us introduce the the free energy:
\BEA
\label{romul}
F(J,h,T)=-T {\sum}_{\by}\bp (\by) \ln {\sum}_{\bx} e^{-\beta H(\by,\bx; J,h)  },
\EEA
defined with the Ising Hamiltonian (\ref{jershalaim}). The entropy
$\Theta$ is expressed via the free energy as [see (\ref{campo_pretorius}, \ref{palatin})]:
\BEA
\label{gambetta}
\Theta =-\partial_T F|_{T\to 0}.
\EEA
Furthermore, we define the following relevant characteristics of MAP: 
\begin{gather}
\label{silvio}
 c={\sum}_{\by}\bp (\by) \rho (\bx|\by) 
\frac{1}{N}{\sum}_{k=1}^N
x_{k} x_{k+1}=\frac{1}{N}\partial_J F ,\nonumber\\
\label{berlusconi}
v={\sum}_{\by}\bp (\by) \rho (\bx|\by) \frac{1}{N} {\sum}_{k=1}^N
y_{k} x_k=\frac{1}{N}\partial_h F,
\end{gather}
Here $c$ accounts for the correlations between
neighbouring spins in the estimated sequence, while $v$
measures the overlap between the estimated and the observed
sequences (the average Hamming distance between the two is simply  $1-v$).
In the limiting case of very weak noise, when the magnitude $h$ of the random fields is large [see (\ref{kaskad})],
we have $v\to v_0= 1$ (observation-dominance),  while $c$  is equal
to the corresponding value $c_0$ of the Markov process ${\cal X}$:
\BEA
\label{mali}
c=
c_0={\sum}_{x_1, x_2}x_1x_2\, p_{\rm  st}(x_1)p(x_2|x_1) = 1-2q.
\EEA
And for very strong noise (the probability of error $\epsilon$ is close to $\frac{1}{2}$,
which means $h\to 0$), $v$ nullifies, while $c$ goes to the corresponding values 
calculated over the prior distribution $\bp(\bx)$: $c={\rm sign}(J)$.

\comment{
\begin{gather}
\label{silvio}
 c=-\frac{\partial }{\partial J}\frac{F(J,h)}{N}
 ={\sum}_{\by}\bp (\by) \rho (\bx|\by) 
\frac{1}{N}{\sum}_{k=1}^N
x_{k} x_{k+1} ,\nonumber\\
 v=-\frac{\partial }{\partial h}\frac{F(J,h)}{N}~~~~~~~~~~~~~~~~~~~~~~~~~~~~~~~~~~\\
\label{berlusconi}
={\sum}_{\by}\bp (\by) \frac{1}{{{\cal N}(\by)}} {\sum}_{\alpha=1}^{{\cal N}(\by)}
\frac{1}{N}{\sum}_{k=1}^N y_k \hat{x}_k ^{[\alpha]}(\by),\nonumber
\end{gather}
}

\section{Recursion relation}
\label{recursion}

Let us return to the partition function (\ref{bala})
\BEA
Z(\by)= \sum_{x_1=\pm 1\ldots x_N=\pm 1} e^{\beta J\sum_{k=1}^N x_{k+1}x_k +\beta h \sum_{k=1}^N y_k x_k   }.
\nonumber
\EEA
We apply to $Z(\by)$ to the following transformations \cite{behn}:
\begin{gather}
\sum_{x_2\ldots x_N} e^{\beta J\sum_{k=2}^N x_{k+1}x_k+\beta h\sum_{k=2}^N y_k x_k} 
\sum_{x_1} 
e^{\beta J x_{1}x_2 + \beta hy_1 x_1   } \nonumber
\\
=\sum_{x_2\ldots x_N} e^{\beta J\sum_{k=2}^N x_{k+1}x_k+\beta h\sum_{k=3}^N y_k x_k
+\beta \xi_2x_2 +\beta B(\xi_1)}, \nonumber
\end{gather}
where $\xi_2=hy_2 +A(\xi_1)$, $\xi_1=hy_1$, and where
\begin{gather}
\label{lolo}
A(u)= \frac{1}{2\beta}\ln\frac{\cosh[\beta J+\beta u]}{\cosh[\beta J-\beta u]},\\
B(u)= \frac{1}{2\beta}\ln\left[ {4\cosh[\beta J+\beta u]}{\cosh[\beta J-\beta u]}\right].
\label{bolo}
\end{gather}
Thus, once the first spin is excluded, the field acting on the
second spin changes from $hy_2$ to $hy_2+A(\xi_1)$.
Note the zero-temperature ($\beta\to\infty$) limits ($J>0$)
\BEA
\label{azbel}
A(u)=u\vartheta (J-|u|) +J\vartheta (u-J)-J\vartheta (-u-J),\\
B(u)=J\vartheta (J-|u|) +u\vartheta (u-J)-u\vartheta (-u-J),
\label{burundi}
\EEA
where $\vartheta(x)=0$ for $x<0$ and $\vartheta(x)=1$ for $x>0$.

Repeating the above steps we express the partition function as follows: 
\BEA
\label{barra}
Z(\by)=e^{\beta \sum_{k=1}^N B(\xi_k)},
\EEA
where $\xi_k$ is obtained from the recursion relation
\BEA
\label{bharat}
\xi_k=h\,y_k+A(\xi_{k-1}), \quad k=1,2,\ldots, N, \quad \xi_0=0.
\EEA
This is a random recursion relation, since $y_k$ are random
quantities governed by the probability $p(\by)$. 
Depending on the value of $y_k$, $\xi_{k+1}$ can take values
$h+ A(\xi_{k-1}) $ or $-h+ A(\xi_{k-1}) $. 
Even when $y_k$ assumes a finite number of values, $\xi_k$ from
(\ref{bharat}) can in principle assume an infinite number of values.
Fortunately, for $T\to 0$, due to the special form (\ref{azbel},
\ref{burundi}) of $A(u)$ and $B(u)$, the number of values assumed by
$\xi_k$ is finite (though it can be large).  It is checked by inspection
that the values taken by $\xi_k$ are parametrized as
$\zeta(n_1,n_2)=(n_1 h + n_2 J)$, where $n_1$ is a positive or negative
integer, while $n_2$ can assume only three values $0,\pm 1$. It can also
be seen that the states $\zeta(n_1,0)$ are not recurrent: once $\xi_k$
takes a value with $n_2=\pm 1$ (note that there is a finite probability
for that), it shall never return to the states $\zeta(n_1,0)$. In the
limit $N\gg 1$, we can completely disregard the states $\zeta(n_1,0)$.

Now recall that the process ${\cal Y}$ with probabilities $p(\by)$ is
not Markov. To make it Markov we should enlarge it by adding the random
variable ${\bf z}$; see (\ref{epatage}).  Here we write the realizations
of this auxiliary Markov process ${\cal Z}$ as ${\bf z}$, so as not to
mix them with those of original process $\bx$.  (${\cal
X}$ and ${\cal Z}$ have identical statistical characteristics, but these
are different processes: ${\cal Z}$ is employed merely for making the
composite process Markov.)
Likewise, we make the process with realizations $[\xi,y]$ Markov by
enlarging it to $[\xi,y,z]$. Let us denote this composite Markov
process by ${\cal C}$. Its conditional probabilities read
\BEA
w(\xi , y, z |\xi' , y', z')
= p(z|z') \pi (y|z)\varphi (\xi |\xi',y),
\label{kruger}
\EEA
where $p(z|z')$ and $\pi (y|z)$ refer to the Markov process
${\cal X}$ and the noise, 
while $ \varphi (\xi |\xi',y)$ takes two values $0$ and $1$, depending
on whether the corresponding transition is allowed or not by recursion
(\ref{bharat}). Now the task is to find all possible values of $\xi_k$, and then to
determine $ \varphi (\xi |\xi',y)$. Before turning to this task, we
relate the characteristics of the studied MAP estimation to
the stationary probabilities $\w(\xi,y,z)$ of
the composite Markov process ${\cal C}$. First we get from
$\w(\xi,y,z)$ the stationary probabilities
$\w(\xi)$. Next we return to (\ref{barra}) and to the definition of free energy
(\ref{romul}). Since the composite Markov process ${\cal C}$ 
will be seen to be ergodic, the free energy can be written as \cite{behn}
\BEA
\label{berkut}
-f(J,h)\equiv -{F(J,h)}/{N}={\sum}_{\xi}\w(\xi)B(\xi),
\EEA
where the summation is taken over all possible [for a given range of $(J,h)$] values of $\xi$.
Once $f(J,h)$ is found, we can apply (\ref{silvio}, \ref{berlusconi}).

As for entropy (\ref{gambetta}) we get from (\ref{bolo}, \ref{barra})
\BEA
F(\by)=-\frac{T}{2} {\sum}_{k=1}^N {\sum}_{s=\pm} \ln \left [2 \cosh [\beta (\xi_k + s J)]\right ].
\EEA
In this expression we should now select the terms which survive $T\to 0$ {\it and} $\partial_T$:
\BEA
-\partial_T F
(\by)|_{T\to 0} =
\frac{\ln 2}{2}\,\,\partial_T \left\{ T\,
{\sum}_{k=1}^N \delta (\xi_k \pm J)
\right\}_{T\to 0},\nonumber
\EEA
where $\delta (.)$ is the Kronecker symbol. In the limit $N\gg 1$,
$\frac{1}{N}\sum_{k=1}^N \delta (\xi_k \pm J)$ should -- with probability
one, i.e., for the elements of the typical set $\Omega({\cal Y})$ -- converge
to $\w(\xi=J)+\w(\xi=-J)$, provided that the composite Markov process is
ergodic. We thus get \cite{behn}
\BEA
\theta\equiv {\Theta}/{N}= {\ln 2}[\, \w(J)+\w(-J)\,]/2.
\label{berlaga}
\EEA
The physical meaning of this formula is that the zero-temperature
entropy can be extensive only when the external field $\xi$ acting on
the spin has the same energy $\xi x_k=\pm 1$ as the spin--spin coupling
constant $J$; see (\ref{om}).  If this is the case, then a macroscopic
amount of spins is frustrated, i.e., the factors influencing those spins
compensate each other, so that their sign is not predetermined even at the
zero temperature. 

\subsection{Stationary states of the recursion}

\begin{figure}
\center
\setlength{\unitlength}{0.154mm}
\begin{picture}(200,155)(20,-180)
        \allinethickness{0.254mm}\put(80,-55){\ellipse{60}{60}} 
        \allinethickness{0.254mm}\put(110,-55){\vector(1,0){50}} 
        \allinethickness{0.254mm}\put(190,-55){\ellipse{60}{60}} 
        \allinethickness{0.254mm}\put(85,-150){\ellipse{60}{60}} 
        \allinethickness{0.254mm}\put(190,-150){\ellipse{60}{60}} 
        \put(80,-60){\makebox(0,10)[cc]{\shortstack{$a_1$}}} 
        \put(190,-60){\makebox(0,10)[cc]{\shortstack{$b_1$}}} 
        \put(85,-155){\makebox(0,10)[cc]{\shortstack{$\bar{a}_1$}}} 
        \put(190,-155){\makebox(0,10)[cc]{\shortstack{$\bar{b}_1$}}} 
        \allinethickness{0.254mm}\put(115,-150){\vector(1,0){45}} 
        \allinethickness{0.254mm}\put(165,-70){\vector(-1,-1){60}} 
        \allinethickness{0.254mm}\put(190,-85){\vector(0,-1){35}} 
        \allinethickness{0.254mm}\put(200,-120){\vector(0,1){40}} 
        \allinethickness{0.254mm}\put(165,-130){\vector(-1,1){60}} 
        \allinethickness{0.254mm}\path(50,-50)(20,-50) 
        \allinethickness{0.254mm}\path(20,-50)(20,-65) 
        \allinethickness{0.254mm}\put(20,-65){\vector(1,0){30}} 
        \allinethickness{0.254mm}\path(55,-140)(25,-140) 
        \allinethickness{0.254mm}\path(25,-140)(25,-155) 
        \allinethickness{0.254mm}\put(25,-155){\vector(1,0){30}} 
\end{picture}
\caption{ The transition graph between various
states (\ref{rusht}) for $m=1$; see (\ref{montana}). 
}
\label{f0}
\end{figure}
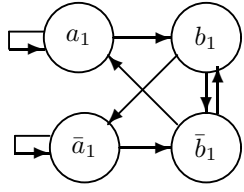

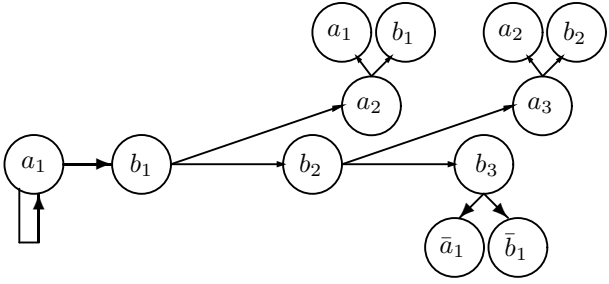
\begin{figure}
\setlength{\unitlength}{0.130mm}
\begin{picture}(615,280)(90,-285)
        \allinethickness{0.254mm}\put(120,-170){\ellipse{60}{60}} 
        \allinethickness{0.254mm}\put(150,-170){\vector(1,0){50}} 
        \allinethickness{0.254mm}\put(230,-170){\ellipse{60}{60}} 
        \allinethickness{0.254mm}\path(105,-195)(105,-250) 
        \allinethickness{0.254mm}\path(105,-250)(125,-250) 
        \allinethickness{0.254mm}\put(125,-250){\vector(0,1){50}} 
        \allinethickness{0.254mm}\put(405,-170){\ellipse{60}{60}} 
        \allinethickness{0.254mm}\put(435,-35){\ellipse{60}{60}} 
        \allinethickness{0.254mm}\put(500,-35){\ellipse{60}{60}} 
        \allinethickness{0.254mm}\put(465,-110){\ellipse{60}{60}} 
        \allinethickness{0.254mm}\path(260,-170)(375,-170)\special{sh 1}\path(375,-170)(369,-168)(369,-170)(369,-172)(375,-170) 
        \allinethickness{0.254mm}\path(260,-170)(435,-110)\special{sh 1}\path(435,-110)(426,-111)(426,-113)(426,-115)(435,-110) 
        \allinethickness{0.254mm}\path(465,-80)(450,-60)\special{sh 1}\path(450,-60)(452,-65)(453,-64)(454,-63)(450,-60) 
        \allinethickness{0.254mm}\path(465,-80)(485,-60)\special{sh 1}\path(485,-60)(480,-63)(481,-64)(482,-65)(485,-60) 
        \allinethickness{0.254mm}\put(580,-170){\ellipse{60}{60}} 
        \allinethickness{0.254mm}\put(550,-255){\ellipse{60}{60}} 
        \allinethickness{0.254mm}\put(615,-255){\ellipse{60}{60}} 
        \allinethickness{0.254mm}\put(610,-35){\ellipse{60}{60}} 
        \allinethickness{0.254mm}\put(675,-35){\ellipse{60}{60}} 
        \allinethickness{0.254mm}\put(640,-110){\ellipse{60}{60}} 
        \allinethickness{0.254mm}\path(435,-170)(550,-170)\special{sh 1}\path(550,-170)(544,-168)(544,-170)(544,-172)(550,-170) 
        \allinethickness{0.254mm}\path(435,-170)(610,-110)\special{sh 1}\path(610,-110)(601,-111)(601,-113)(601,-115)(610,-110) 
        \allinethickness{0.254mm}\path(640,-80)(625,-60)\special{sh 1}\path(625,-60)(627,-65)(628,-64)(629,-63)(625,-60) 
        \allinethickness{0.254mm}\path(640,-80)(660,-60)\special{sh 1}\path(660,-60)(655,-63)(656,-64)(657,-65)(660,-60) 
        \put(120,-175){\makebox(0,10)[cc]{\shortstack{$a_1$}}} 
        \put(230,-175){\makebox(0,10)[cc]{\shortstack{$b_1$}}} 
        \put(548,-260){\makebox(0,10)[cc]{\shortstack{$\bar{a}_1$}}} 
        \put(613,-260){\makebox(0,10)[cc]{\shortstack{$\bar{b}_1$}}} 
        \put(403,-175){\makebox(0,10)[cc]{\shortstack{$b_2$}}} 
        \put(463,-115){\makebox(0,10)[cc]{\shortstack{$a_2$}}} 
        \put(433,-40){\makebox(0,10)[cc]{\shortstack{$a_1$}}} 
        \put(498,-40){\makebox(0,10)[cc]{\shortstack{$b_1$}}} 
        \put(608,-40){\makebox(0,10)[cc]{\shortstack{$a_2$}}} 
        \put(673,-40){\makebox(0,10)[cc]{\shortstack{$b_2$}}} 
        \put(638,-115){\makebox(0,10)[cc]{\shortstack{$a_3$}}} 
        \put(583,-175){\makebox(0,10)[cc]{\shortstack{$b_3$}}} 
        \allinethickness{0.254mm}\put(580,-200){\vector(-1,-1){25}} 
        \allinethickness{0.254mm}\put(580,-200){\vector(1,-1){25}} 
\end{picture}
\caption{ Transitions between various
states (\ref{rusht}) for $m=3$; 
see (\ref{montana}). This
is one half of the real transition graph. The second half is obtained
from the above one by adding bars to all above symbols: $a\to \bar{a}$,
$b\to \bar{b}$; see (\ref{khan}, \ref{rusht}).
}
\label{f1}
\end{figure}

For given $J$ and $h$ define an integer $m$ as
\BEA
\label{montana}
{2J}/{(m-1)}\,>\, h\, >\, {2J}/{m}, \quad m=1,2,\ldots
\EEA
Note that the case $h>2J$ (and there is no upper limit on $h$) 
corresponds to $m=1$. One can check that for each integer $m$
the recurrent states $[\xi,y]$ assumed by the recursion (\ref{bharat})
can be parametrized as
\begin{gather}
\label{khan}
\{a_i,\,  b_i, \, \bar{a}_i, \, \bar{b}_i\}_{i=1}^m, \\
 a_i=[(2-i)h+J,1]\equiv [\alpha_i,1], ~~~~
   \bar{a}_i\equiv [\,-\alpha_i,-1\,],\nonumber\\
 b_i=[-ih+J,-1]\,\equiv[\beta_i,-1], ~~~~
  \bar{b}_i\equiv [-\beta_i,1].
\label{rusht}
\end{gather}
Note the symmetry $\bar{a}_i=-a_i$ and $\bar{b}_i=-b_i$.
The transitions between these states|which via the binary function $\varphi$ determine 
the transition matrix in (\ref{kruger})|are illustrated in Figs.~\ref{f0} and \ref{f1} 
for $m=1$ and $m=3$, respectively. 
The reader can easily generalize the latter graph to an arbitrary $m$.

We are now prepared to write down from (\ref{kruger}, \ref{khan}, \ref{rusht}) and Fig.~\ref{f0} the following 
transition matrix for the composite Markov process ${\cal C}$ with $m=1$
\begin{eqnarray}
W=    
\left[
     \begin{array}{cc|cc}
      w_{a_1|a_1} & 0 & 0 & w_{a_1|\bar{b}_1}  \\
      w_{b_1|a_1} & 0 & 0 & w_{b_1|\bar{b}_1}  \\
    \hline
      0          & w_{\bar{a}_1|b_1} & w_{\bar{a}_1|\bar{a}_1} & 0 \\
      0          & w_{\bar{b}_1|b_1} & w_{\bar{b}_1|\bar{a}_1} & 0 \\
     \end{array}   
    \right].
\label{dede}
  \end{eqnarray}
This is a block matrix composed of $2\times 2$ matrices
(hence the actual size of $W$ in (\ref{dede}) is $8\times 8$): 
$0$ means the $2\times 2$ matrix with
all its elements equal to $0$, and 
\BEA
\label{brento1}
w_{...|a_1}=w_{...|\bar{b}_1}=P, ~~~ w_{...|b_1}=w_{...|\bar{a}_1}=M,\\
P_{xx'}\equiv \pi (+1|x) p(x|x'), \,\,\,
M_{xx'} \equiv \pi (-1|x) p(x|x'),
\EEA
where $x,x'=\pm 1$. Note that $P+M$ is equal to the transition matrix of the Markov process ${\cal X}$.
Once the $8\times 1$ stationary probability vector $w$ of $W$ is found
from $Ww=w$, we get $\w(\alpha_1)=w_1+w_2$,
$\w(\beta_1)=w_3+w_4$, $\w(\bar{\alpha}_1)=w_5+w_6$, and $\w(\bar{\beta}_1)=w_7+w_8$.

For a general $m$ the following $m\times m$ matrices serve as building blocks for the matrix $W$
\BEA
L=\{ L_{i,i+1}=1, i=1,\ldots,m-1  \}, \, E=\{ E_{1,1}=1  \}, \nonumber\\
U=\{ U_{i+1,i}=1, i=1,\ldots,m-1  \},\,\
S=\{ S_{1,m}=1  \},\nonumber
\EEA
where all not indicated elements are equal to zero. The transition matrix for a general $m$ reads
[see Figs. \ref{f0}, \ref{f1}]
\begin{gather}
\label{comrad}
W =
\left[\begin{array}{rr}
M & 0 \\
P & 0 \\
\end{array}\right]\otimes 
\left[\begin{array}{rr}
0 & 0 \\
0 & E \\
\end{array}\right] ~~~~~~~~~~~~~~~~~~~~~~~\\
+
\left[\begin{array}{rr}
0 & P \\
0 & M \\
\end{array}\right]\otimes 
\left[\begin{array}{rr}
L & 0 \\
0 & 0 \\
\end{array}\right]
\,+\,
\left[\begin{array}{rr}
0 & M \\
0 & P \\
\end{array}\right]\otimes 
\left[\begin{array}{rr}
0 & 0 \\
S & L \\
\end{array}\right]\nonumber\\
+ \left[\begin{array}{rr}
P & 0 \\
M & 0 \\
\end{array}\right]\otimes 
\left[\begin{array}{rr}
U & 0 \\
0 & 0 \\
\end{array}\right]
\,+\,
\left[\begin{array}{rr}
M & 0 \\
P & 0 \\
\end{array}\right]\otimes 
\left[\begin{array}{rr}
0 & 0 \\
0 & U \\
\end{array}\right]\nonumber\\
+ \left[\begin{array}{rr}
P & 0 \\
M & 0 \\
\end{array}\right]\otimes 
\left[\begin{array}{rr}
E & 0 \\
0 & 0 \\
\end{array}\right]
\,+\,
\left[\begin{array}{rr}
0 & P \\
0 & M \\
\end{array}\right]\otimes 
\left[\begin{array}{rr}
0 & S \\
0 & 0 \\
\end{array}\right].\nonumber
\end{gather}
The left matrices of each tensor
product is a block matrix; each block consists of one $2\times 2$
matrix. The right matrices of each tensor product are also block matrices;
now each block consists of one $m\times m$ matrix. The zero
$m\times m$ matrix is written as $0$. The overall size of $W$ is $8m\times 8m$, since each state
in (\ref{khan}) is augmented by two realizations of the hidden Markov process. 

Note that going from one value of $m$ to another amounts to
changing the dimension of the matrices $E$, $U$, $S$ and $L$. Since
these matrices are sparse, efficient numerical algorithms of treating
them are available, even for larger values of $m$.

\section{MAP inference}
\label{mapo}
Let us indicate how the quantities of interest are expressed via the
stationary probability $\w$ of the Markov process ${\cal C}$ (obtained
from (\ref{comrad})). 
Recall that since the estimated process is
unbiased, we are interested in the second moment $c$, overlap $v$ and
entropy $\theta$. The former two quantities have to be obtained via
the free energy. To this end, we trace out the redundant
variables in the stationary probability of $W$ to obtain the following probabilities ($i=1,\ldots, m$):
\begin{gather}
\w_m (\alpha_i)=\w_m (\bar{\alpha}_i ), \quad \w_m (\beta_k)=
\w_m (\bar{\beta}_k  ),
\label{ord}
\end{gather}
where the equalities in (\ref{ord}) are due to the symmetry of the unbiased situation.
We add a lower index to relevant quantities (e.g., to
$\w$'s) to indicate the specific value of $m$. 
Recall that, e.g., $\w_1(\alpha_1)$ and $\w_2(\alpha_1)$ are in general different quantities,
since they belong to different Markov processes ${\cal C}_1$ and ${\cal C}_2$, respectively.

Due to (\ref{ord}), we shall need only the probabilities $\w_m(\alpha_k)$ and $\w_m(\beta_k)$
that normalize to one-half:
\BEA
\label{ecuador}
{\sum}_{k=1}^m \left[\, \w_m(\alpha_k)+ \w_m(\beta_k)\, \right]={1}/{2}.
\EEA
The free energy then reads (see (\ref{burundi}, \ref{berkut}))
\BEA
-\frac{f_m}{2}=
{\sum}_{k=1}^m \left[\, \w_m(\alpha_k)B(\alpha_k)+ \w_m(\beta_k)B(\beta_k)\, \right]\nonumber\\
      = h[\, \w_m(\alpha_1)+m \w_m (\beta_m)   \,] +J[\,\half -2 \w_m(\beta_m)   \, ].
\label{pareto}
\EEA
Now we make use of the fact that free energy is a continuous function of its parameters~\footnote{Outside phase 
transitions free energy is smooth, while at the phase-transition
points it has to be at least continuous, since, besides
being the generating function for calculating various averages,  free
energy is also a measure of dynamic stability, and at the phase-transition points both
phases are equally stable by definition (see~\cite{landau} for more details).}, which in  our case  implies
\BEA
f_m=f_{m+1} \quad {\rm at} \quad h={2J}/{m}, \quad m=1,2,\ldots
\EEA
This leads from (\ref{pareto}) to
\BEA
\w_m(\alpha_1)= \w_{m+1}(\alpha_1)+\w_{m+1}(\beta_m).
\label{bolivar}
\EEA
One can confirm (\ref{bolivar}) from (\ref{babylon1}, \ref{babylon2}, \ref{elam}).
Note that (\ref{bolivar}) will hold for all values of $\e$ and $q$, since it
does not depend on $h$ and/or $J$ (the formalism holds
without requiring any specific relation between $h,\, J$ and $\e,\, q$).
Combining (\ref{pareto}) with  (\ref{berlusconi}) from Section~\ref{silvio}, 
we obtain for the second moment $c_m$ of the estimated sequence and 
the overlap $v_m$
\BEA
\label{togo}
c_m= 1-4\w_m(\beta_m),\,\,
v_m= 2\w_m(\alpha_1)+2m \w_m(\beta_m).
\EEA
As seen from (\ref{khan}, \ref{rusht}), if the relations
(\ref{montana}) hold [recall that they are strict inequalities], there
are only two realizations $\alpha_2=J$ and $\bar{\alpha}_2=-J$,
which, according to (\ref{berlaga}), contribute into the entropy.
Recalling also (\ref{ord}), we get ($m=1,2,\ldots$)
\BEA
\label{mongol}
\theta =  \w_m(\alpha_2)\, \ln 2, 
\quad {\rm for }\quad \frac{2J}{m-1}>h>\frac{2J}{m}.
\EEA
This relation holds for $m=1$, if we assume $\w_1(\alpha_2)=0$.  

At the transition points
$h=\frac{2J}{m}$ between the various regimes (\ref{montana}), there are
more states that contribute into the entropy. The reader can verify
that
\BEA
\label{shuudan}
\theta =  [\w_m(\alpha_2)+\w_m(\beta_{m})]\, \ln 2, \quad {\rm if }\quad h={2J}/{m}.
\EEA
This equation is written down assuming that the value of $\theta$ at $h=\frac{2J}{m}$ does not depend on whether
the latter point is reached as $h\to\frac{2J}{m}+0$ or as $h\to\frac{2J}{m}-0$. This assumptions leads from
(\ref{shuudan}) to:
\BEA
\label{inner_mongolia}
\w_m(\alpha_2)+\w_m(\beta_{m})=\w_{m+1}(\alpha_2)+\w_{m+1}(\beta_{m}). 
\EEA
This relation has the same origin as the continuity of 
the free energy.

\subsection{The regime $m=1$ or $h>2J$.}

We deduce for the stationary probabilities from (\ref{comrad})
\BEA
\label{babylon1}
\w_1(\alpha_1)\equiv \w_1(h+J)
= \frac{1-q}{2}+\epsilon(1-\epsilon)(2q-1),\\
\w_1(\beta_1)\equiv \w_1(-h+J)=
\frac{q}{2}-\epsilon(1-\epsilon)(2q-1).
\label{babylon2}
\EEA
This implies from (\ref{togo}) $v_1=1$, $c_1=(1-2q)(1-2\epsilon)^2$.
This is in fact the Maximum Likelihood (ML) regime: the noise is so
small (or $h$ is so large) that the estimation is completely governed by
the observations: $v_1=1$. The second moment $c_1$ of the estimated
sequence in this regime is given by the original value $c_0=1-2q$ (see
(\ref{mali})) times the squared error probability $(1-2\e)^2$. The
entropy in this regime is zero (see (\ref{mongol})): $\theta_1=0$. In
this sense the estimation is uniquely determined by observations. We
stress that the ML and MAP schemes agree with each other not only for
very small, but also for finite noises.

\subsection{The regimes $m=2$ and $m=3$.}

For more compact presentation of the probabilities, let us introduce
separate notations for the noise strength and the Markov correlator
$g=\e(1-\e)$, $u=1-2q$, where $0<g<\oo$ and $0<u<1$; see (\ref{mali}).
The probabilities obtained from (\ref{comrad}) are written as
\begin{gather}
\w_2(\alpha_1)= \frac{ \o+  \left(\frac{1}{2}-3 g\right) u   }{  3 - (1 + 2 g) u }, \,\,\,\,
\w_2(\beta_1)= \frac{\o+\left(g-\frac{1}{2}\right) u }{ 3 - (1 + 2 g) u }       ,\nonumber\\
\w_2(\alpha_2)= \frac{\oo+\frac{1}{2} (3 g-1) u+ \frac{1}{4} (1-2 g (4 g+1)) u^2   }{ 3 - (1 + 2 g) u },\nonumber\\
\w_2(\beta_2)= \frac{\oo-\frac{g u}{2}+ \frac{1}{4} \left(8 g^2+2 g-1\right) u^2  }{ 3 - (1 + 2 g) u}.
\label{elam}
\end{gather}
We skip a tedious analytical expressions for $\w_3$.  

\begin{figure}[!htb]
\centering
\subfigure[]{
    \includegraphics[width=7cm]{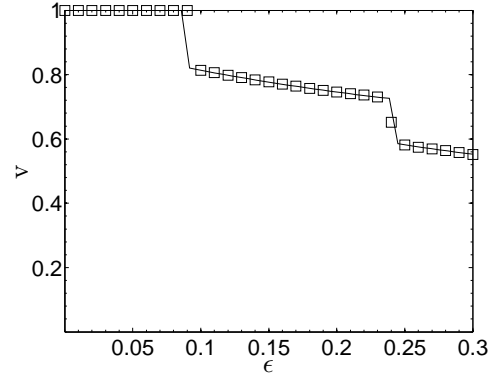} \label{fig_v}
    } 
    \subfigure[]{
    \includegraphics[width=7cm]{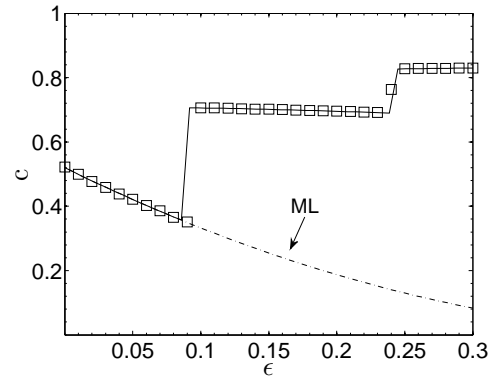} \label{fig_c}
    }
    \subfigure[]{
    \includegraphics[width=7cm]{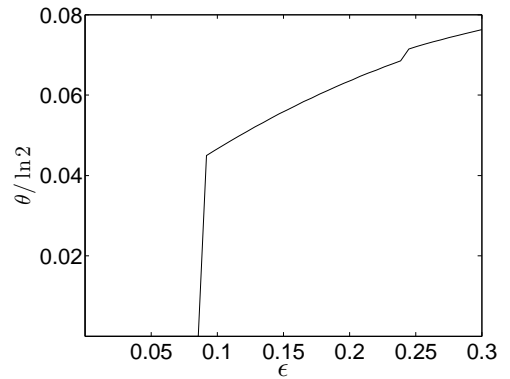} \label{fig_entropy}
    }
     \caption{MAP characteristics versus the noise intensity in the regimes $m=1,2,3$ for $q=0.24$: (a) Overlap $v$; (b)  the second moment $c$; (c) Entropy $\frac{\theta}{\ln2}$. In (a) and (b) the open squares represent simulation results, obtained by running the Viterbi algorithm and calculating  the respective quantities directly. We used sequences of size $10^4$, and averaged the results over $100$ random trials.}
     \label{fig_0}
\end{figure}

The values of $c$ and $v$ deduced from (\ref{togo}, \ref{elam}) are shown in
Figs. \ref{fig_v} and \ref{fig_c}. We compare those values with the results obtained by actually finding the MAP estimate via the Viterbi algorithm, and calculating those quantities directly. It is seen that at the regime change
points $h=2J$ and $h=J$, $v$ and $c$ experience sudden jumps, or
first-order phase transitions. Remarkably, those features are perfectly reproduced in the simulations, as shown in Figure~\ref{fig_v} and \ref{fig_c}. For instance, in   the ML regime $h>2J$ ($0<\e<0.09068$), the overlap 
$v=1$ indicating that the estimation is governed solely by observations.  At $h=2J$ it jumps sharply, and then monotonically decreases  in the regime $2J>h>J$.  More generally,  $v$ decays, both monotonously and via jumps, towards the prior-dominated value $v=0$. 

Consider the second moment $c$ of the estimated sequence shown in
Figure~\ref{fig_c}. We see that  $c$  is nearly a constant for each given $m\geq 2$. This is the main
virtue of MAP scheme as compared to the ML scheme: While the latter
predicts a $c$ that quickly decays with the noise as
$c_{ML}=(1-2q)(1-2\e)^2$ (the dotted line in the plot), the proper MAP
value of $c$ is not far from its noise-free value $c_0=1-2q$, and is
nearly a constant for a finite range of noise strength $\e$. This
advantage of MAP over ML is due to supporting the estimation process by
the priors. Indeed, the values of the overlap indicate that the
estimated sequence is not completely driven by the observations, though
it is still not very far from them. Upon increasing $\e$ towards its
maximal value $\e=\frac{1}{2}$, $c$ experiences jumps during each regime
change. For larger $m$ these jumps are smaller and more frequent,
leading $c$ to its prior-dominated value $1$. 

\begin{table}
\begin{tabular}{|c|c|c|c|c|}
\hline
$m$ &    $4$  & $5$ & $6$ & $7$   \\
\hline
 $\e$  & $0.3700$ & $0.3910$  & $0.4100$  & $0.421$   \\
\hline
$\theta/ \ln 2$&   0.07308       & 0.06587         & 0.05925    & 0.05349    \\
\hline
\end{tabular}
\caption{Regular values of entropy $\frac{\theta}{\ln 2}$ for $q=0.24$; see (\ref{mongol}).
This table continues Fig. \ref{fig_entropy} towards larger values of the noise strength $\e$.
}
\label{tab_1}
\end{table}

Now let us focus on the entropy: It naturally nullifies in the ML regime ($\theta_1=0$), while in the regime $m=2$  the
entropy $\theta_2$ is monotonously increasing with
$\e$, as shown in Figure \ref{fig_entropy}. At $h=2J$ (the phase transition point, where $m$ changes from
1 to 2) $\theta_2$ experiences a jump, which is again usual for
first-order phase-transitions. $\theta$ maximizes at an intermediate
value of $\e$, and then decays to zero for $\e\to\frac{1}{2}$, see Table
\ref{tab_1}; at this point the present approach reduces to a
ferromagnetic 1d Ising model without magnetic fields. This model has a
trivial ground-state structure and hence zero entropy. We also note that right  at the transition points $h=\frac{2J}{m}$ the values of $\theta$
is different; see Table \ref{tab_2}. The largest value is attained for
$h=2J$.
\begin{table}
\begin{tabular}{|c|c|c|c|c|c|}
\hline
$h$ &    $2J$  & $J$  &  $2J/3$   &  $J/2$   &  $2J/5$    \\
\hline
 $\e$  & $0.0907$ & $0.2400$  & $0.3598$  & $0.3867$ & 0.4051  \\
\hline
$\theta/\ln 2$&   0.1629       & 0.1462         & 0.1220    & 0.0992  & 0.0831         \\
\hline
\end{tabular}
\caption{The special values of entropy $\frac{\theta}{\ln 2}$ for $q=0.24$.
}

\label{tab_2}
\end{table}

\begin{figure}[!h]
\vspace{0.2cm}
\center
\includegraphics[width=8cm]{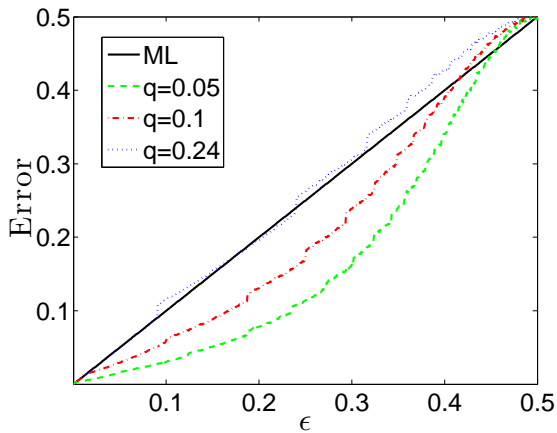}
\caption{The average error rate given by the normalized Hamming distance between the true and the estimated sequences.}
\hfill
\label{fig_3}
\end{figure}

Finally, we would like to note that the second moment of the estimated
sequence, $c$, is an indirect measure of accuracy.  In practice, one is
restricted to use such indirect measures as  information about the
true sequence might not available.  In Figure~\ref{fig_3} we present the average error rate for 
the MAP estimation, which is given by the normalized Hamming distance
between the true and Viterbi--decoded sequences, plotted against the noise
intensity. Also shown is the average error rate of ML estimation,
which is simply $\epsilon$. For vanishing noise, both
MAP and ML yield the same average error. Upon increasing the noise intensity, the MAP estimation error  behaves differently depending on the parameter $q$: For small
values of $q$, MAP is always superior to ML for  a wide range 
of noise intensities. For larger values of $q$, however, the situation is more
complicated: Although both methods perform similarly, there
are some differences and crossovers between the two at intermediate
noise intensities, as shown in Figure~\ref{fig_3}. \comment{It is also seen that  for  high noise intensities ($\epsilon\approx1/2$) MAP estimation  error is slightly worse compared to ML, although both methods produce rather poor}
\vspace{-8mm}

\section{Discussion}

We theoretically examined Maximum a Posteriori (MAP)
estimation for hidden Markov sequences, and found that MAP yields 
 a non--trivial solution structure even for the simple binary and
unbiased hidden Markov process considered here. We demonstrated that for a finite range 
of noise intensities, there is no difference between
 MAP and  Maximum Likelihood (ML) estimations, as the solution is
observation--dominated. While it was expected that the two methods agree
for a vanishing noise, {\em the fact of their exact agreement for a finite range of the
noise is non-trivial}. Furthermore,  upon increasing the noise
intensity the MAP solution switches between different operational
regimes that are separated by first--order phase transitions. In
particular, a first-order phase-transition separates the regime where
MAP and ML agree exactly. At this transition point the influence 
of the prior information becomes comparable to the influence of observations.
In the vicinity of the first-order phase-transitions the performance of MAP (e.g., characteristics of the
estimated sequence) changes abruptly. This means that a small change in
the noise intensity may lead to a large change in the performance. In other
words, the phase-transition points should be avoided in applications.

 For practical applications of HMM (e.g., in speech recognition, or machine translation)
it is not enough to know the single solution that provides the largest
posteriori probability \cite{foreman}. At the very least, one should also
know how many sequences have a posteriori probabilities sufficiently
close to the optimal one. Motivated by this fact, we studied the number
${\cal N}$ of MAP-solutions that have (for long sequences $N\to \infty$)
almost equal logarithms of the posterior probability. A finite
$\theta=\frac{1}{N}\ln {\cal N}$ means that there is an {\it exponential}
number of solutions with posterior probabilities slightly less than the 
optimal. We found that $\theta$ is finite whenever MAP differs from ML.  We believe that this theoretical result might have practical implications as well. For instance, in applications such as statistical machine translation, one usually considers  top $K$ solutions to the inference problem, and then chooses one according to some heuristics. Our result suggests that one needs to be careful with this practice  whenever $\theta$ is non--zero, as one might  discard a large number of nearly optimal solutions if $K$ is not chosen sufficiently large.  

We also note that our work is directly related to the notion of {\em
trackability}, which can be intuitively defined as one's ability to
(accurately) track certain stochastic processes~\cite{Crespi2008,sheng2005}. In fact, a similar
phase--transition in the number of solutions was reported by Crespi {et.
al.}~\cite{Crespi2008} for so called {\em weak models}, where the
entries in the HMP transition and emission matrices are either $0$ or
$1$. For more general stochastic processes, an information-theoretical
characterization of trackability was suggested in \cite{sheng2005}.
Within this approach, the accuracy is characterized by the probability
${\rm Pr}[\hat{\bx}\not=\bx]$ of the estimated sequence $\hat{\bx}$ not
being equal to the actual one, while the structure of the solution space
is described via the number of elements $|\Omega|$ in the (conditional)
typical set $\Omega$ of $\bx$ sequences given an observed sequence $\by$
(complexity). Both these quantities relate to the conditional entropy
$-{\sum}_{\bx,\by}{\rm Pr}(\bx,\by)\ln{\rm Pr}(\bx|\by)$.  We 
note that whereas the accuracy and the complexity measures of
\cite{sheng2005} deteriorate even for a small (but generic) noise
intensity, our approach of defining trackability in terms of the
zero--temperature entropy of the Ising Hamiltonian
(Equation~\ref{jershalaim}) suggests that a process can be trackable in the MAP sense even in the
presence of moderate noise. 

Finally, we would like to note that another interesting feature of the
MAP estimation is that its characteristics ($c$ and $v$) change only slightly  in between the phase-transition points. In  contrast to ML estimation, which deteriorates (at least linearly) when increasing the noise, 
 MAP estimation is stable for a finite range of noise intensities. Thus, 
although  MAP estimation may be less accurate compared
to ML, it can be still useful as far as its stability is concerned,
provided that its range of application is selected carefully.

There are several directions for further developments. First, we intend
to obtain analytical results for the average error rate to complement
the empirical analysis presented in Figure~\ref{fig_3}. Furthermore, one
can think of a semi--supervised MAP estimation, where one has
(possibly noisy) knowledge about the states of the hidden process at
particular times. Remarkably, the framework presented here allows a
natural generalization to this case. Indeed, one simply needs to modify
the Ising energy function by adding quenched fields at the corresponding
locations in the chain. Finally, it will be interesting to generalize
the analysis presented here beyond the binary hidden Markov processes
considered here. In this case, the MAP optimization problem can be
mapped to a Potts model. We would like to note that the behavior observed in the simple binary model can be explained by 
the emergence of a finite fraction of ``frustrated" spins, where the frustration can be attributed 
to two competing tendencies -- accommodating observations on one hand, and the 
hidden (Markovian) dynamical model on the other. Since this mechanism is rather general, we  believe that most features of the MAP scheme uncovered here via an exact analysis of the simplest binary model will survive in more general situations. 
\subsubsection*{Acknowledgements} 

Armen Allahverdyan thanks USC Information Sciences Institute for hospitality and support. This research was supported by the U.S. ARO MURI grant W911NF--06--1--0094.

\end{document}